\documentclass{article}
\usepackage{ijcai24}
\usepackage{times}          
\usepackage{soul}
\usepackage{url}
\usepackage[hidelinks]{hyperref}
\usepackage[utf8]{inputenc}
\usepackage[small]{caption}
\usepackage[switch]{lineno}
\usepackage[utf8]{inputenc} 
\usepackage[T1]{fontenc}    
\usepackage{booktabs}       
\usepackage{amsfonts}       
\usepackage{nicefrac}       
\usepackage{microtype}      
\usepackage{xcolor}         
\usepackage{graphicx}       
\usepackage{upquote}        
\usepackage{tikz}           
\usepackage{amssymb,amsmath,amsthm,mathtools}   
\usepackage{algorithm, algorithmic}            

\newtheorem{lemma}{Lemma}
\newtheorem{definition}{Definition}

\author{
Abbas Mehrabian$^1$\and
Ankit Anand$^1$\and
Hyunjik Kim$^1$\and
Nicolas Sonnerat$^1$\and
Matej Balog$^1$\and \\ 
Gheorghe Comanici$^1$\and
Tudor Berariu$^2$\and
Andrew Lee$^1$\and
Anian Ruoss$^1$\and
Anna Bulanova$^1$\and \\
Daniel Toyama$^1$\and
Sam Blackwell$^1$\and
Bernardino Romera Paredes$^1$\and
Petar Veli\v{c}kovi\'{c}$^1$\and \\
Laurent Orseau$^1$\and
Joonkyung Lee$^3$\and
Anurag Murty Naredla$^4$\and
Doina Precup$^1$\And\\
Adam Zsolt Wagner$^5$\\
\affiliations
$^1$Google DeepMind\\
$^2$Imperial College London\\
$^3$Yonsei University\\
$^4$University of Bonn\\
$^5$Worcester Polytechnic Institute \\
}

\hypersetup{ 
    colorlinks=true,
    linkcolor=teal,
    filecolor=red,
    urlcolor=teal,
    citecolor=teal,
    pdftitle={Finding Increasingly Large Extremal Graphs with AlphaZero and Tabu Search},
    }
\newcommand{\score}{\operatorname{s}}
\newcommand{\flip}{\bigoplus}

\title{Finding Increasingly Large Extremal Graphs \\ with AlphaZero and Tabu Search\thanks{To appear in the proceedings of IJCAI 2024, the 33rd International Joint
Conference on Artificial Intelligence.}}

\begin{document}

\maketitle

\begin{abstract}
This work proposes a new learning-to-search benchmark and uses AI to discover new mathematical knowledge related to an open conjecture of Erd{\H{o}}s (1975) in extremal graph theory. The problem is to find graphs with a given size (number of nodes) that maximize the number of edges without having 3- or 4-cycles. We formulate this as a sequential decision-making problem and compare AlphaZero, a neural network-guided tree search, with tabu search, a heuristic local search method. Using either method, by introducing a curriculum---jump-starting the search for larger graphs using good graphs found at smaller sizes---we improve the state-of-the-art lower bounds for several sizes. We also propose a flexible graph-generation environment and a permutation-invariant network architecture for learning to search in the space of graphs.
\end{abstract}

\section{Introduction}
With the recent advances in neural networks, artificial intelligence (AI) methods have achieved tremendous success in multiple domains like game playing~\cite{alphazero}, biology~\cite{alphafold}, mathematics~\cite{davies2021}, and robotics~\cite{peng&al18}. Mathematics is of particular interest to AI researchers due to its challenging multistep reasoning structure, open-ended problems, and limited data. While automated theorem proving has always been of interest to AI researchers as a reasoning benchmark~\cite{abdelaziz2022trail,aygun22proving,kovacs2013first,lample2022hypertree,polu2020generative,schulz2002brainiac}, some recent work have used machine learning to solve research problems across the fields of representation theory, knot theory, graph theory, and matrix algebra~\cite{davies2021,alphatensor,wagner_paper}.

Many mathematical problems can be modeled as searching for an object or a structure of desired characteristics in an extremely large space. Indeed, automated theorem proving is often modeled as searching for a sequence of operations---a proof---in an ever-growing space of operands with a few operators. Another example is counterexample generation~\cite{wagner_paper}, where the object of interest is a counterexample to a particular conjecture or a mathematical construction that improves the bounds for a problem. The recent work of AlphaTensor~\cite{alphatensor} also relies on neural network-guided tree search to find novel tensor decompositions that result in faster matrix multiplication algorithms.

Inspired by these, we focus on a classical extremal graph theory problem, studied by~\cite{erdos_conjecture}, which is to find, for any given number of nodes, a graph that maximizes its number of edges but is constrained not to have a 3-cycle or a 4-cycle. While the problem is simple to state, mathematicians have not found optimal constructions for all sizes: the maximum number of edges is known for up to 53 nodes~\cite{exact_values}, and lower bounds using local search have been reported for up to 200 nodes~\cite{bong,garnick1993extremal}. Because strong local-search methods have been developed for this problem, it provides a challenging benchmark for learning-based search methods. We believe that developing new methods for graph problems could inspire methods for related fields such as drug discovery and chip design, where the goal is to find a graph object minimizing or maximizing a given objective function. Discovering new optimal solutions to this problem could lead to more efficient designs in other problems, such as data center organization and optimization as well as game theory.

In this paper, we use reinforcement learning (RL) and formulate graph generation as a sequential decision making process. In contrast to the graph-generation RL environment used by~\cite{wagner_paper}, which starts from an empty graph and adds edges one by one in a fixed order, we start from an arbitrary graph and add/remove edges in an arbitrary order. This RL environment, called the \emph{edge-flipping} environment, has at least two advantages: (a) we can start from known ``good'' graphs to find even better graphs, and (b) we can scale to larger sizes by using a curriculum of starting from slightly smaller graphs. As our RL agent, we use the state-of-the-art AlphaZero~\cite{alphago_zero} algorithm, which has shown impressive success in a variety of domains such as tensor decomposition~\cite{alphatensor}, discovering new sorting algorithms~\cite{alphadev}, and game playing~\cite{alphago_zero}. 

\paragraph{A novel neural network architecture.}
Since AlphaZero is guided by a neural network, we also need a representation that aligns with the invariants of the search space. We deal with simple undirected graphs, so graph neural networks naturally come to mind~\cite{velivckovic2023}; but we go beyond them and introduce a novel representation, the \emph{Pairformer}: unlike traditional graph neural networks, which pass messages between nodes, Pairformers pass messages between \emph{pairs of nodes}. Pairformers burden us with additional computational cost but are significantly better at detecting cycles. 

The Pairformer has edge features for all existing and non-existing edges between nodes, whereas standard GNNs have edge features only for existing edges in the graph. This property of Pairformers allows the network to directly reason about non-existing edges, which is important for the policy network to understand which edges should be added or removed in the graph. Combined with triangle self-attention updates, this enables effective processing of neighboring edge features that can capture presence or absence of cycles with only few Pairformer layers.
The Pairformer network architecture is novel and provides significant improvements over the ResNet architecture. We hope that this architecture will be useful to tackle other graph problems, too.

\paragraph{Incremental learning.}
When searching over all graphs, one challenge is that the number of graphs with a given number of nodes $n$ increases exponentially with $n$; thus, finding optimal graphs becomes significantly harder as $n$ grows. Interestingly, known optimal graphs for this problem have a \emph{substructure property}: in many cases, optimal graphs of a given size are near-subgraphs of optimal graphs of larger sizes (see, e.g., \cite[Theorem 3]{backelin}). Thus, finding near-optimal graphs for smaller $n$ can serve as a stepping stone to find good graphs for larger values of $n$.
This property can be used to construct a curriculum: start from discovered graphs of a given size, generate novel solutions of a larger size, and repeat. Our edge-flipping environment provides the flexibility to start from any graph and add or drop edges arbitrarily, so we are not restricted to supergraphs of the starting graph and can reach all graphs of that size. Deploying these ideas in AlphaZero, we develop \emph{Incremental AlphaZero} and improve the lower bounds for sizes 64 to 136.

This way of scaling to larger sizes is related to the idea of curriculum learning~\cite{cl_survey}, a widely-used method in RL, especially for solving hard exploration problems in many domains, including robotics~\cite{rubikscube} and
automated theorem proving~\cite{aygun22proving} 
as well as solving the Rubik's cube and other difficult puzzles~\cite{agostinelli2019rubik,orseau2023ltscm}. Note that the term ``curriculum learning'' has been used with different meanings in machine learning literature; in this paper, by \emph{curriculum} we mean solving the problem on the smaller size first and then using the solution of the smaller problem to solve the same problem on the larger size.

\paragraph{Incremental local search.}
The substructure property can enhance other types of search as well. We develop an incremental version of tabu search, a known local search method~\cite{tabu_search}, where the initial graph for each size is sampled from a previously-discovered ``good'' graph of a smaller size. This algorithm also improves over the state of the art. 
Our ablation shows that both search strategies improve significantly by this idea of incremental learning where we scale to larger sizes by using high scoring graphs of slightly smaller size.

\paragraph{Summary of contributions.}
We introduce a challenging benchmark for learning-to-search in large state spaces, inspired by an open problem in extremal graph theory, whose best solutions thus far are achieved by local search. We formulate graph generation as an edge-flipping RL environment and introduce the novel representation Pairformer, which is well-suited for detecting cycles in undirected graphs. We introduce the  idea of incremental search (curriculum) to local search methods as well as AlphaZero, and show that kickstarting from solutions of smaller size is a key ingredient for improving the results on this extremal graph theory problem. We improve the lower bounds for the problem for all graph sizes from 64 to 134, and we release these graphs to the research community to aid further research: \url{https://storage.googleapis.com/gdm_girth5_graphs/girth5_graphs.zip}. 

\section{Problem Description}\label{sec:background}
A \emph{$k$-cycle} is a cycle with $k$ nodes. Let $G$ be a simple, undirected $n$-node graph that has no 3-cycles. What is the maximum number of edges that $G$ can have? Mantel~\cite{mantel} proved that the answer is precisely $\lfloor n^2/4 \rfloor$, initiating the field of extremal graph theory.
Tur\'an~\cite{turan} generalized this result to cliques and found, for any $k$, the maximum number of edges that an $n$-node graph without $k$-cliques can have.

Generally, for a set $\mathcal H$ of graphs, let $\operatorname{ex}(n, \mathcal H)$ denote the maximum number of edges in an $n$-node graph that does not contain any member of $\mathcal H$ as a subgraph (the symbol $\operatorname{ex}$ stands for ``extremal''). Calculating $\operatorname{ex}(n, \mathcal H)$ for various graph classes $\mathcal H$ is a central problem in extremal graph theory. In this paper, we study
\begin{equation}\label{def_fn}
f(n) \coloneqq \operatorname{ex}(n, \{C_3, C_4\}).
\end{equation}
See Table~\ref{small_fn} in appendix for values of $f(n)$ for small $n$. 
We know $\operatorname{ex}(n, \{C_3\}) = \lfloor n^2/4 \rfloor$ by Mantel's theorem and 
$\lim_{n\to\infty}\operatorname{ex}(n, \{C_4\})/(n\sqrt{n}) = 1/2$ by~\cite{brown,erdosrenyisos}, but no formula has been found for 
$f(n)$, and even its asymptotic behavior is not understood well.

\cite{erdos_conjecture} conjectured that 
\(\lim_{n\to\infty} \frac{f(n)}{n\sqrt{n}} = \frac{1}{2\sqrt 2}\). This conjecture has remained open since 1975.
The tightest bounds are due to~\cite{garnick1993extremal}, who proved
\begin{equation}\label{garnicketlbounds}
\frac{1}{2\sqrt 2} \leq \lim_{n\to\infty} \frac{f(n)}{n\sqrt{n}} \leq \frac{1}{2}.
\end{equation}

Motivated by this conjecture, we want to estimate the value of $f(n)$ for specific values of $n$. It is known that $f(n) \leq n\sqrt{n-1}/2$ for all $n$~\cite{garnick1993extremal}.
The exact value of $f(n)$ is known when $n\leq53$ \cite{exact_values}, and constructive lower bounds have been reported for all $n\leq 200$~\cite{bong,garnick1993extremal}.
Our goal is to improve these lower bounds for $54\leq n \leq 200$.
Hence, we want to find $n$-node graphs without 3-cycles or 4-cycles that have as many edges as possible.

The \emph{size} of a graph is its number of nodes. For any graph $G$, we denote its number of edges, 3-cycles, and 4-cycles by $e(G)$, $\triangle(G)$, and $\square(G)$, respectively. We say that a graph $G$ is \emph{feasible} if it has no 3-cycles and no 4-cycles.
The \emph{score} of a graph $G$ is defined as
\begin{equation*}
\score(G) \coloneqq e(G) - \triangle(G) - \square(G).
\end{equation*}
The following lemma (proof in Appendix~\ref{sec:proof}) implies that, for any given number of nodes $n$, proving lower bounds for $f(n)$ is equivalent to maximizing the score over all $n$-node graphs.
\begin{lemma}\label{lemma:score}
For any $n$-node graph $G$, we have $\score(G) \leq f(n)$;
and there exists at least one $n$-node feasible graph for which equality holds.
\end{lemma}

In light of Lemma~\ref{lemma:score}, we can formulate the problem of maximizing $f(n)$ in two ways: we can maximize $e(G)$ over \emph{feasible} $n$-node graphs or maximize $\score(G)$ over \emph{all} $n$-node graphs. The two formulations have the same optimal value, but their search space differs. The first formulation has a smaller search space, but the second one, which we use, allows us to define a convenient \emph{neighborhood function}, which helps our algorithms navigate the space of graphs more smoothly.

\begin{definition}[$\flip$, flipping]\label{def:flipping}
Let $G$ be a graph and let $u$ and $v$ be two of its nodes.
If $uv$ is an edge in $G$, then $G \flip uv$ is obtained by removing the edge $uv$ from $G$;
otherwise, $G \flip uv$ is obtained by adding the edge $uv$.
In either case, we say $G \flip uv$ is obtained by \emph{flipping} $uv$.
\end{definition}

The flipping operation has two desirable properties:
first, any $n$-node graph assumes exactly $\binom{n}{2}$ flips, a technical convenience for RL agents' action space;
second, any $n$-node graph can be reached from any other $n$-node graph by doing up to $\binom{n}{2}$ many flips; that is, there are no ``dead ends.''

\section{Graph Generation as an RL Environment}
We define \emph{graph generation} as a sequential decision making process, where we start from an $n$-node graph $G$ and, at each step, modify it by adding or removing an edge $e$ to obtain $G'= G \flip e$. More formally, graph generation is a deterministic finite-horizon Markov Decision Process (MDP) $\mathcal{M} = \{\mathcal{S}, \mathcal{A}, R, \mathcal{T}, H\}$, where
the state space, $\mathcal{S}$, consists of all simple undirected graphs of size $n$;
the action space, $\mathcal{A} \coloneqq \{ (i, j): 1 \leq i < j \leq n\}$, consists of all edges of the complete graph of size $n$;
the one-step reward function (defined below) is $R: \mathcal{S} \times \mathcal{A} \to \mathbb{R}$;
the deterministic transition function, $\mathcal{T}$, is defined as $\mathcal{T}(G, e) \coloneqq G \flip e$;
and the horizon, $H$, denotes the number of steps in each episode.

We define the reward function as $R(G, e) \coloneqq \score(G \flip e) - \score(G)$, i.e., the reward equals the change in the score after taking the action. We call this \emph{the telescopic reward}, as the rewards accumulated over time form a telescoping series, making the episode return equal to $\score(G_H) - \score(G_0)$. Note that $\score(G_H)$ is precisely the objective value that we want to maximize.
(Often in RL, a discount factor is introduced, and subsequent rewards are discounted when computing the return; but we do not introduce a discount factor here, as then the return would have been different from the actual objective function.)
In our experiments, we found that the telescoping reward performs much better than the \emph{non-telescoping} reward, where the reward is given at the end of each episode and equals $\score(G_H)$.

This \emph{edge-flipping environment} provides more flexibility than environments in which the graph is built, for instance, by deciding about the edges one by one in a fixed order~\cite{wagner_paper}. An advantage of the edge-flipping environment is that every action is reversible: in this MDP, any $n$-node graph can be reached from any other $n$-node graph. This property is particularly useful when using a curriculum, as one can use \emph{state resetting}~\cite{florensa17a,hosu2016playing,salimans2018learning} to warm up exploration from high-scoring initial graphs; e.g., start from high quality graphs of size $n-k$ (garnished with $k$ isolated nodes) to build the desired graph of size $n$. On the other hand, reversibility can cause learning instability: since there is no termination action, an agent could end up flipping one of the edges indefinitely. To avoid such issues, we set a fixed horizon length.

\section{AlphaZero for Graph Generation}
\label{sec:alphazerodetails}
AlphaZero is a reinforcement learning algorithm that demonstrated superhuman performance on Go, through self-play, without using any human knowledge~\cite{alphago_zero}. It was then adapted to show superhuman performance in other games such as chess and shogi~\cite{alphazero}. Recently, a version of AlphaZero was adapted to find faster algorithms for matrix multiplication. This new model, named AlphaTensor, improved Strassen's matrix-multiplication algorithm for some sizes for the first time after 50 years~\cite{alphatensor}. The AlphaZero algorithm combines Monte Carlo Tree Search (MCTS), a heuristic search algorithm, with deep neural networks to represent the state space, e.g., a chessboard position. In our application of AlphaZero, each state is a simple undirected graph and each action is adding or removing an edge.

The edge-flipping environment is a deterministic MDP, where both the transition matrix and the reward function are fully known and deterministic, thus most search and planning algorithms are applicable. MCTS builds a finite tree rooted at the current state and, based on the statistics gathered from the neighboring states, selects the next action. Many successful works using MCTS use some variant of the upper confidence bound rule~\cite{kocsis&szepezvari06} to balance exploration and exploitation when expanding the tree. While traditional approaches used Monte Carlo rollouts to estimate the value of a leaf state, in the last decade this has largely been replaced by a neural network, called the \emph{value network}. 
Another neural network, called the \emph{policy network}, determines which child to expand next. Often, the policy and value networks share the same first few layers. (They have the same latent representation, or torso, but they have different heads.) In AlphaZero, both the policy and value networks are trained using previously observed trajectories---see~\cite{alphazero} for details. 

For updating the value of a state---which is a node in the MCTS tree---standard MCTS expands the node and uses the average value of the children. Since we want to maximize the best-case return rather than the expected return, it may appear more suitable to use the maximum value of the children to update the value of the node. We  attempted this approach but it did not yield improvements.

One common issue with AlphaZero is encouraging it to diversely explore the space of possible trajectories. We attempted a few ideas to diversify, such as increasing UCB exploration parameter and also for each trajectory, if same graph ($s_i$) is encountered which has been seen in the previous timesteps ($t<i$) within the trajectory, we discourage this behavior by giving a small negative reward; but none of these approaches improved the result on top of starting from good graphs of smaller size.

\subsection{Network Representation}
\label{sec:network_rep_new}
To find a good representation for this problem of avoiding short cycles, we tested different architectures on the supervised learning problem of cycle detection, using this as a proxy for our RL problem for fast experiment turnaround. In particular, we compared ResNets \cite{resnet2016}, Pointer Graph Networks~\cite{velivckovic2020pointer}, Graph Attention Networks~\cite{velivckovic2017graph}, and a novel architecture called the \emph{Pairformer}, described below. We studied node and edge level binary classification tasks (whether a node or an edge is part of a short cycle) as well as graph level tasks (whether a graph contains a short cycle). The Pairformer gave the best performance, hence this is the architecture we used in the RL setting.

An intuitive understanding for the Pairformer can be gained by recognising its main difference with standard GNNs: the Pairformer has edge features for all existing and non-existing edges between nodes, whereas standard GNNs have edge features only for existing edges in the graph. This property of Pairformers allows the network to directly reason about non-existing edges, which is important for the policy network to understand which edges should be added or removed in the edge-flipping environment. Combined with triangle self-attention updates (explained below), this enables effective processing of neighboring edge features that can capture presence or absence of cycles with only few Pairformer layers. We found that the additional computational cost is worth it.

The Pairformer is a simplified version of Evoformer, used in AlphaFold~\cite{alphafold}. Each Evoformer block has two branches of computation: one processes the multiple sequence alignment (MSA) representation and the other one processes the pair representation. The Pairformer only uses the pair representation branch, which processes per-edge features and has shape $(n, n, c)$. We set $c=64$ in our implementations. Within the pair representation branch, each Pairformer block is composed of triangle self-attention blocks (row-wise multihead self-attention followed by column-wise multihead self-attention) followed by fully-connected layers with LayerNorm~\cite{ba2016layer}. We omitted the triangle multiplicative updates in the original Evoformer as they had minimal effect on performance for our tasks. A key difference with standard graph neural networks is that instead of only having features for existing edges, the Pairformer has features for all $\binom{n}{2}$ pairs of nodes, whether they correspond to existing edges or not. We believe that considering non-existing edges is crucial for the Pairformer to inform the policy for deciding whether to add new edges to the graph or not. This architecture is used as the torso network, which receives the current graph as input and outputs a representation that is consumed by the policy and value heads.

The current graph is given input as an $n \times n$ adjacency matrix. Since we use a single network for multiple sizes, we condition the torso and the policy head on the graph size $n$ by concatenating each input with a matrix of 1s on the principal $m \times m$ (where $m < n$) submatrix and 0s everywhere else (concatenate along the channel dimension). This lets us use a shared set of parameters for multiple graph sizes without a separate network for each size.

A good model architecture should not only be expressive but also have fast inference in order for acting to be fast enough to quickly generate lots of data for the learner to optimize the model. The downside of the Pairformer is its $O(n^3)$ runtime, while ResNet's runtime is $O(n^2)$. Hence there exists a trade-off between expressiveness and speed, and we experimentally found that combining a small Pairformer torso with a larger ResNet policy head provides the best balance. Using a ResNet for the torso performs much worse, implying that the expressiveness that the Pairformer brings to the torso's representation of the input graph is indispensable---see Figure~\ref{fig:rn_pf}. For the value head, we used a feed-forward network over the representation provided by the Pairformer.

Another important detail is that although the environment supports only $\binom{n}{2}$ many actions, the last layer of our policy network has twice as many logits: for each edge, there is one logit for adding that edge and another logit for removing that edge. This means half of the logits correspond to invalid actions (e.g., adding an edge for an existing edge). We mask these invalid actions so a valid probability distribution is induced on the valid set of $\binom{n}{2}$ actions.

\subsection{Distributed Implementation and Joint Learning Across Multiple Sizes}
\label{sec:multiple-size}
We use a distributed implementation of AlphaZero with multiple processing units: in each run, there are multiple actors, one replay buffer, and one learner. Each actor has a copy of the networks (supplied by the learner) and generates episodes, which are inserted into the replay buffer. The learner repeatedly samples an episode from the replay buffer to update the policy and value networks. The policy network is trained using the cross-entropy loss, where the ground truth label is assumed to be the decision taken at the root of the MCTS tree. The value network is trained using regression on the future return (sum of the future rewards) at each state of an episode.

For efficiency and transfer-learning across multiple sizes, we jointly train a single network for multiple sizes: each run of AlphaZero is provided with a list of target sizes, and each actor samples a target size uniformly at random from this list. The network input is modified in this case by padding it by 0s to turn it into a $target \times target$ matrix, while appending another plane to the observation, each entry of whose principal $size \times size$ submatrix is 1 and the rest are 0. This helps to run experiments for multiple sizes jointly and ensures transfer-learning across different sizes. 

\subsection{Incremental AlphaZero}
\label{sec:inc_az}
A key observation about our problem is that, in many cases, the optimal graph for a given size is nearly a subgraph of an optimal graph for a larger size. For example, by \cite[Theorem 3]{backelin}, all the optimal graphs for sizes 40, 45, 47, 48, 49 are subgraphs of the optimal graph for size 50.
While this is not strictly true for all the sizes, it can be used as a heuristic to guide the search.  
As all the optimal graphs up to size 52 are known~\cite{mckay_graphs}, we use high-scoring graphs of smaller sizes (garnished with a suitable number of isolated nodes) as the initial graph and iterate. Hence if the results for smaller sizes improve, this hopefully leads to subsequent improvements for larger sizes as well. This not only exploits the approximate substructure property for the problem but is also related the well-known idea of curriculum learning in machine learning~\cite{bengio2009curriculum}.

Specifically, starting episodes from the empty graph leads to a difficult credit assignment problem for the RL agent, as the horizon should be long. Instead, we start from a high-scoring graph of size $n-k$ to build the target graph of size $n$. We observe that this choice of initial graph is critical for the performance of AlphaZero. It leads to a shorter horizon and more effective credit assignment in each episode. We call the resulting algorithm as \emph{Incremental AlphaZero}.

\section{Tabu Search for Graph Generation}
\label{sec:tabusearchdetails}
Tabu search is a well-known iterative local search method~\cite{tabu_search}:
given an objective function and a neighborhood structure over a set of states, it repeatedly moves from the current state to the neighboring state with the highest objective value, until some stopping condition is met. To avoid getting stuck at local minima, tabu search bans revisiting recently-visited states---hence the name ``tabu'' search.

In our case, the states are the graphs of a given size, the graphs obtained by flipping a single edge are the neighbors of the current graph, and the objective function is $\score(G) = e(G) - \triangle(G) - \square(G).$ Our tabu search algorithm (Algorithm~\ref{alg:tabu_search}) slightly differs from the typical definition; instead of banning visiting \emph{states} that were recently visited, we ban playing the recently-played \emph{actions}. Namely, we ban re-flipping edges that were flipped recently. This idea, inspired by~\cite{ramseymultiplicity}, results in a slightly faster algorithm than the usual tabu search.
Note that the algorithm needs an initial graph $G_0$---we will describe later how it's chosen---and has a single hyperparameter: the history size, denoted by $h$ in Algorithm~\ref{alg:tabu_search}, which determines the number of iterations flipping an edge is banned once it is flipped. Recall that  $\flip$ denotes flipping an edge.
\begin{algorithm}
\caption{Our version of tabu search}\label{alg:tabu_search}
\begin{algorithmic}[1]
\REQUIRE $G_0$ is an $n$-node graph with nodes indexed from 1 to $n$,
and $0\leq h < \binom{n}{2}$
\ENSURE $BestGraph$ is the highest-scoring graph found during search
\STATE $Tabu \gets$ a first-in-first-out queue of fixed size $h$
\STATE $Actions \gets \{ (i, j): 1 \leq i < j \leq n\}$
\STATE $BestGraph \gets G_0$
\FOR{$i \gets 1,2,\dots, iterations$}
\STATE $ValidActions \gets Actions \setminus Tabu$
\STATE $BestActions \gets \arg\max_e \{\score(G_{i-1} \flip e) : e \in ValidActions\}$
\STATE $Action \gets$ random action chosen from $BestActions$
\STATE $G_i \gets G_{i-1} \flip Action$
\STATE Insert $Action$ into $Tabu$
\IF{$\score(G_i)>\score(BestGraph)$}
    \STATE $BestGraph \gets G_i$
\ENDIF
\ENDFOR
\end{algorithmic}
\end{algorithm}
\subsection{Incremental Tabu Search}
The \emph{incremental} tabu search algorithm is inspired by the idea mentioned in section \ref{sec:inc_az}: we let the tabu search at each size start its search from one of the best graphs found at smaller sizes. Say we want to find lower bounds for $f(n)$ for some range $n \in \{a, \dots, b\}$. Incremental tabu search is a distributed algorithm with $b-a+1$ parallel workers (processing units), indexed from $a$ to $b$, where the worker with index $n$ searches for graphs of size $n$. The workers need a common memory to share the graphs they have found: suppose that $BestGraphs[n]$, for $n \in \{a, \dots, b\}$, is a set of graphs that all workers have access to. (We can initialize it to contain just the empty graph of size $n$.) The algorithm for the size-$n$ worker appears in Algorithm~\ref{alg:incremental_tabu_search}. 
\begin{algorithm}
\caption{Incremental tabu search (worker for size $n$)}\label{alg:incremental_tabu_search}
\begin{algorithmic}[1]
\REQUIRE $0\leq K$ and $a \leq n \leq b$ and $BestGraphs[n]$ is a set of graphs of size $n$
\ENSURE $BestGraphs[n]$ contains the set of highest-score graphs found during the search
\WHILE{True}
\STATE Sample $k$ randomly from  $\{1, \dots, K\}$
\STATE Sample $G_0$ randomly from $BestGraphs[n-k]$
\STATE Add $k$ isolated nodes to $G_0$
\STATE Run tabu search starting from $G_0$
\STATE $BestFoundGraph \gets $ best graph found by tabu search
\STATE Choose $ExistingGraph$ arbitrarily from $BestGraphs[n]$
\IF{$\score(BestFoundGraph) > \score(ExistingGraph)$}
    \STATE $BestGraphs[n] \gets \{ BestFoundGraph \}$
\ELSE
    \STATE Add $BestFoundGraph$ to $BestGraphs[n]$
\ENDIF
\ENDWHILE
\end{algorithmic}
\end{algorithm}

\pagebreak

\section{Experiments and Results}
We compare five methods: tabu search starting from the empty graph; incremental tabu search, which uses a curriculum to use high-scoring graphs found at each size as the starting point for larger sizes; AlphaZero starting from the empty graph; incremental AlphaZero, which uses a curriculum;
and Wagner's cross-entropy method~\cite{wagner_paper}, the first machine-learning method to find counterexamples for mathematical conjecture. For AlphaZero, we also perform ablations on the choice of network representation. Since $f(n) = \Theta(n\sqrt n)$ (see~\eqref{garnicketlbounds}) we have normalized the scores by $n\sqrt n$ in the plots. The hyperparameters are provided in appendix.

\begin{figure*}[t]
\centering
\includegraphics[width=0.70\textwidth]{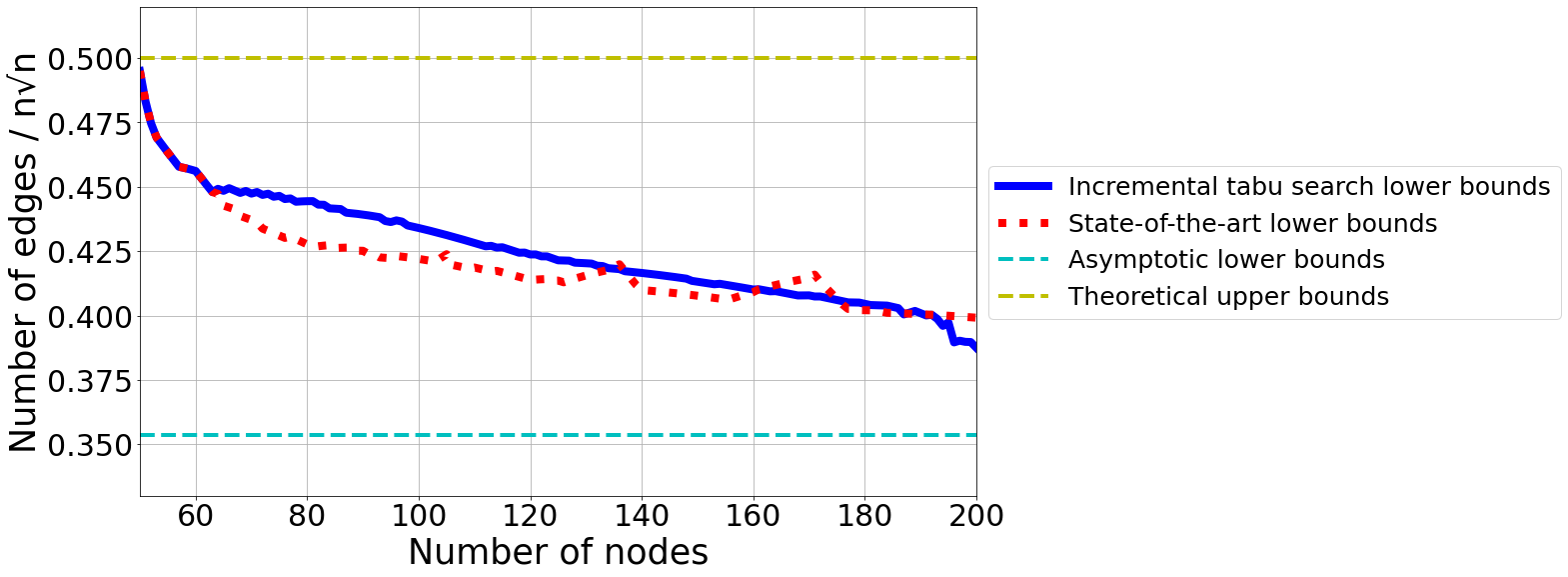}
\caption{Normalized scores, given by $\frac{\text{number of edges}}{n\sqrt n}$, are plotted versus size, $n$. AlphaZero with curriculum (not plotted) achieves the same score as incremental tabu search for 41 of the sizes from 54 to 100. Erd{\H{o}}s conjectured that both the red and blue curves converge to the cyan horizontal line as $n\to\infty$.}
\label{fig:sota}
\end{figure*}

\begin{figure}[t]
\centering
\includegraphics[width=0.23\textwidth]{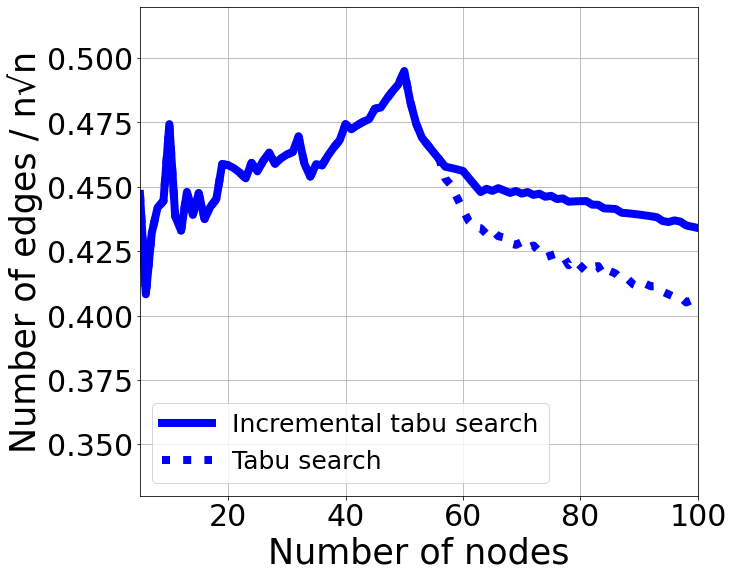}
\includegraphics[width=0.23\textwidth]{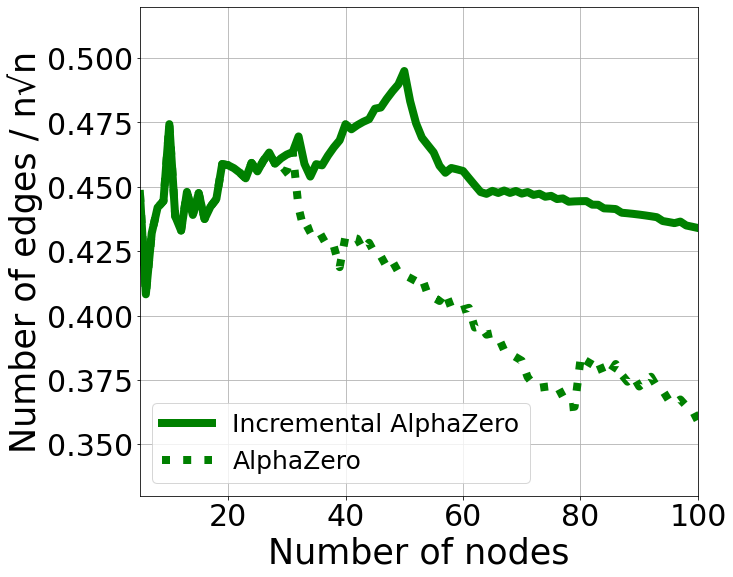}
\caption{Left: Incremental tabu search, which uses a curriculum, performs increasingly better than tabu search without curriculum, for larger problem sizes.
Right: Adding a curriculum improves the performance of AlphaZero significantly, especially on larger sizes.}\label{fig:curriculum}
\end{figure}

For each size $n$, we started the episodes in incremental AlphaZero with one of the high-scoring graphs found by incremental tabu search at size $n-k$, where $k$ is chosen randomly between 1 and 4. (We did this for technical convenience, but we believe similar results are achievable if we sample from graphs of smaller sizes generated by previous runs of AlphaZero.)

A key hyperparameter is the episode length (the horizon). An overly long episode length is not only wasteful but also hinders learning, as the agent may find an optimal graph in the middle of an episode but still has to flip edges to reach the end of the episode. On the other hand, an overly short episode length would hinder the exploration as the agent is limited to the vicinity of the initial graph. We ran AlphaZero (without incremental search) for sizes 5 to 100, split these sizes in five nearly-equal buckets of 5--20, 21--40, 41--60, 61--80, and 81--100, and set horizons to 80, 160, 240, 320, and 434, respectively. With incremental search, since we start from a good graph of smaller size, we can choose shorter horizons; we experimented with horizon lengths of 30, 50 and 100 but found the results don't change much beyond 30.

For tabu search, we tried various history sizes but size 5 worked best. For each size, we ran 32 parallel copies of tabu search for seven days, restarting every 1000 iterations and merging the results. For incremental tabu search, we initialized $BestGraphs[n]$ (see Algorithm~\ref{alg:incremental_tabu_search}) to contain the set of graphs published by~\cite{mckay_graphs} (for $n=1,2,\dots, 64$), set the history size to 5, and the $K$ in incremental tabu search to 4---it is important this is greater than 1.

\begin{figure}[t]
\centering
\includegraphics[width = 0.28\textwidth]{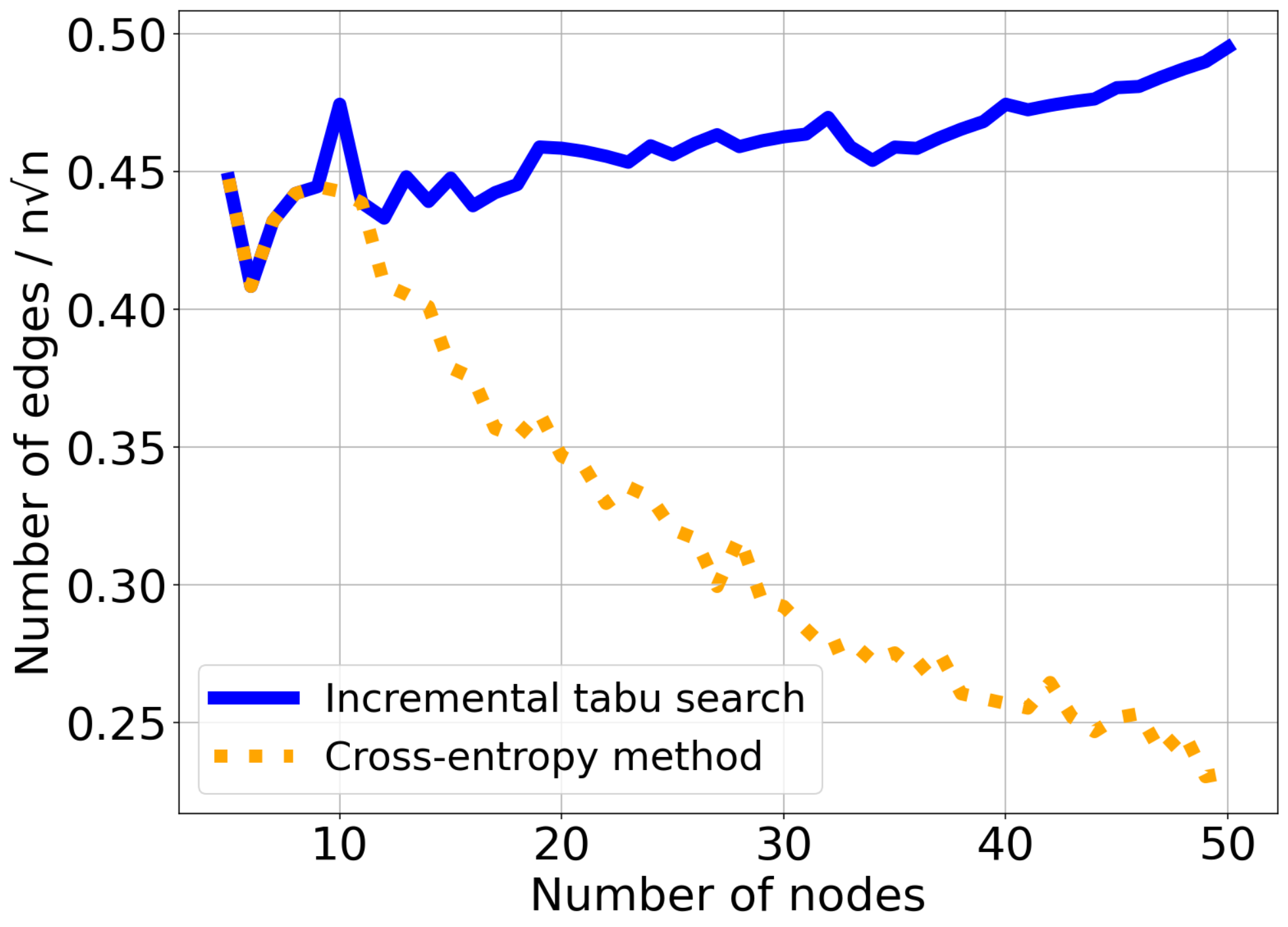}
\caption{Incremental tabu search versus the cross-entropy method.}
\label{fig:its_ce}
\end{figure}

\paragraph{Comparison with the state-of-the-art lower bounds.} 
As Figure~\ref{fig:sota} shows, incremental tabu search improves the state-of-the-art lower bounds\footnote{
In Figure~\ref{fig:sota}, state-of-the-art lower bounds are from \cite{garnick1993extremal,abajo,garnick_email,mckay_graphs} and theoretical upper bounds are from~\cite{garnick1993extremal}.
We have not compared against the lower bounds reported in~\cite{bong}, as neither the graphs achieving those bounds nor the method for generating them are presented in~\cite{bong}; still, incremental tabu search improves over the lower bounds reported in~\cite{bong} for $n\in\{64, \dots, 76\}$.} when $n \in \{64,\dots,134\} \cup \{138,\dots,160\} \cup \{176,\dots,186\}\cup\{188,\dots,190\}$. For a concrete example, see the appendix, where we have also listed the lower bounds achieved by incremental tabu search for $n=1,2,\dots, 200$. Incremental AlphaZero also improves over the state-of-the-art lower bounds on many sizes, and is exactly on par with incremental tabu search on all sizes between 54 to 100, except $n=$ 56, 57, 64, 66, 77, and 96, where incremental tabu search leads by one edge.  We observe that Wagner's cross entropy method~\cite{wagner_paper} performs much worse than both tabu search and AlphaZero---see Figure~\ref{fig:its_ce}.

\begin{figure}[t]
\centering
\includegraphics[width=0.23\textwidth]{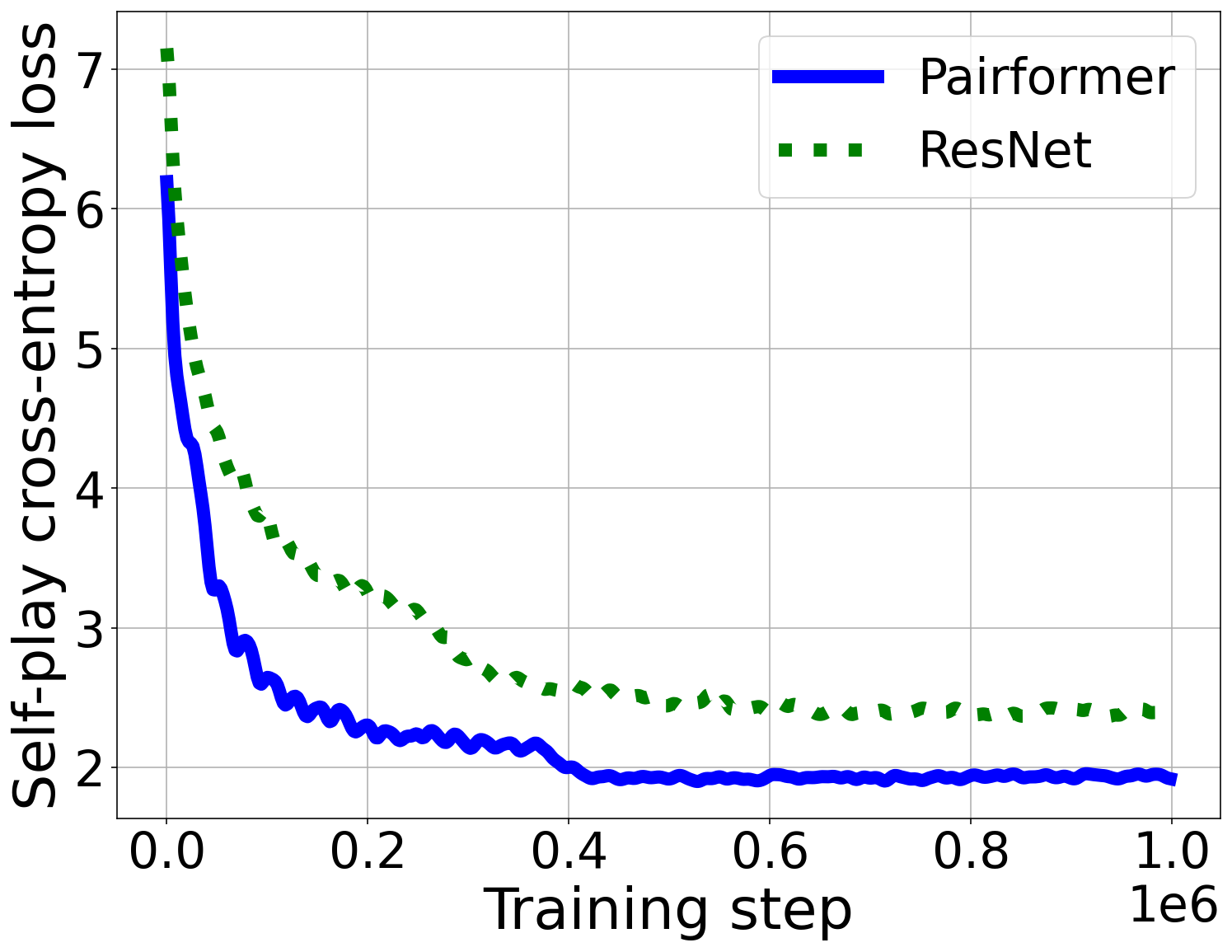}
\includegraphics[width=0.23\textwidth]{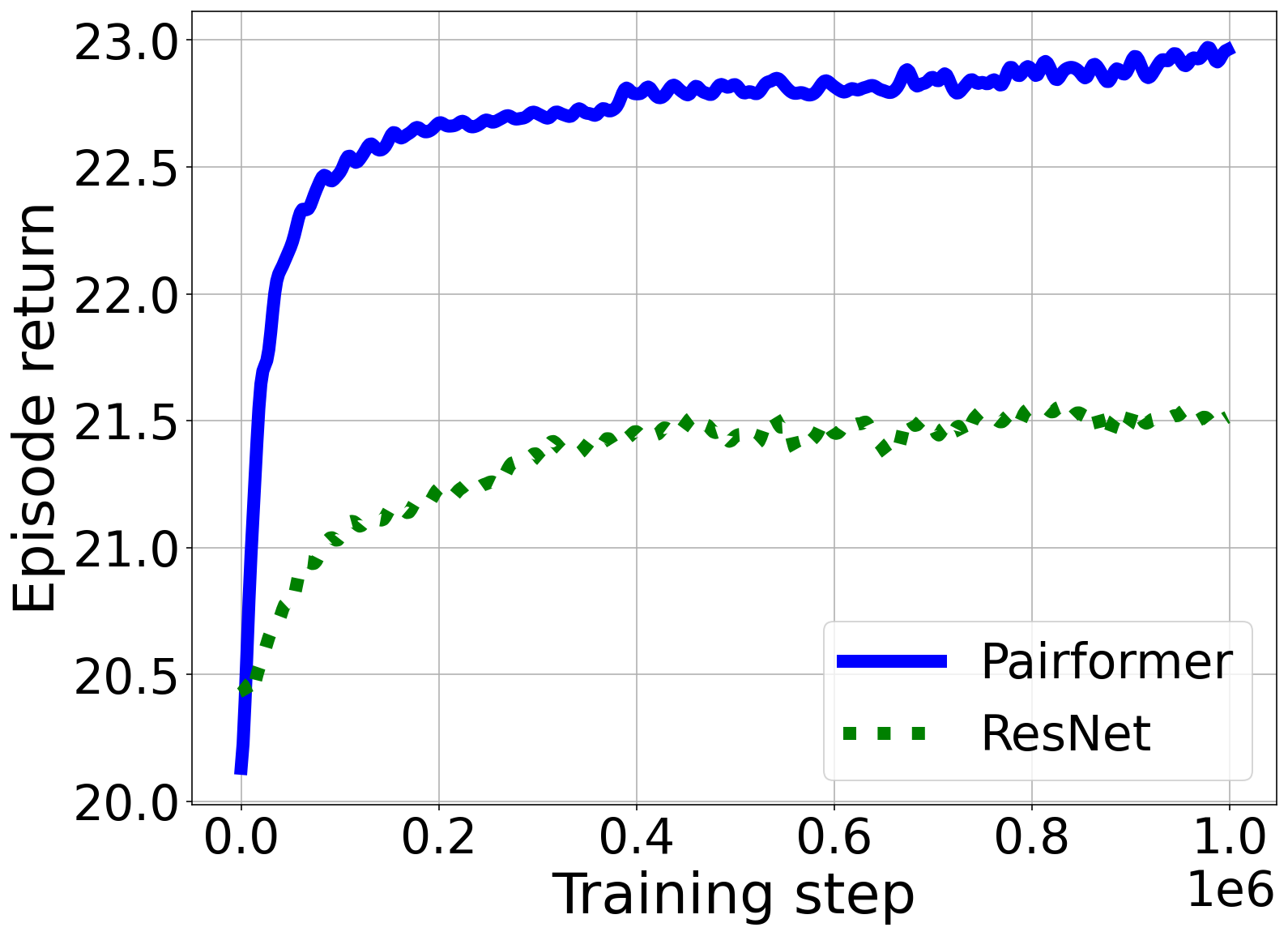}
\caption{Left: The policy cross-entropy loss of Pairformer and ResNet during online training of AlphaZero  on joint training for graph sizes [80, 100] with curriculum. Pairformer minimizes the loss faster as it captures invariances and other graph structures. Right: Average episode return of Pairformer and ResNet during training of AlphaZero using the edge-flipping environment on joint training for graph sizes [80, 100] with curriculum. In both plots, the average is taken over 3 seeds  and Gaussian smoothing with $\sigma=2$ is applied.} \label{fig:rn_pf}
\end{figure}

\paragraph{Benefits of using incremental search} Without curriculum and incremental search, the agent must build graphs of a given size starting from the empty graph, while with incremental search, the agent starts from a previously-found graph of a smaller size and flips some edges to obtain a graph of the desired size. Figure~\ref{fig:curriculum} (left) illustrates, for sizes 54 to 100, the benefit of using a curriculum for tabu search, which increases significantly as the number of nodes increases. Figure~\ref{fig:curriculum} (right) shows a similar plot for AlphaZero: AlphaZero matches Incremental AlphaZero up to size around 30 but deteriorates afterwards. We believe the reasons are large episode lengths and the difficulties of exploration and credit assignment. We conclude that using incremental learning (and curriculum) is a vital ingredient for applying RL to this problem and presumably for any optimization problem with a substructure property and a huge state space.

\paragraph{Comparing representations in AlphaZero.}
\label{sec:comparingrepresentations}
We compare our novel representation, Pairformer, with ResNet~\cite{resnet2016}, which has been extensively used in literature, especially in environments where the observation is a matrix or an image. Since a graph can be naturally expressed as an adjacency matrix, we compare the ResNet architecture, which is oblivious to the graph structure, with the Pairformer architecture. We use a ResNet with 10 layers and 256 output channels. Figure \ref{fig:rn_pf} (left) shows the policy's cross-entropy loss during training, and Figure \ref{fig:rn_pf} (right) shows the average episode return. For this experiment, we focus on joint training for graph sizes [80, 100] with Incremental AlphaZero. We observe that Pairformer performs better on both metrics. Nevertheless, the final scores obtained by ResNet and Pairformer are equal except on sizes 80, 86 and 93, where Pairformer leads by one score point. It should be noted that the above experiments are for parameters sizes (number of layers, attention heads) beyond which we didn't see a performance improvement for either representation and the exact number of parameters may differ for both representations.

\paragraph{Comparing the cross-entropy method with incremental tabu search.}
\label{sec:cross-entropy}
We compare incremental tabu search with the cross-entropy method~\cite{wagner_paper} in Figure~\ref{fig:its_ce}. (The hyperparameters are provided in Appendix~\ref{sec:hyperparams}.) We observe that incremental tabu search outperforms cross-entropy method by a big margin as the size of the graph grows. (Incremental AlphaZero performs similarly to incremental tabu search, so we haven't plotted it.)

While it would have been useful to compare the resources used by each method, the different nature of the algorithms hinders a fair comparison, especially because some are sequential and others are parallel. AlphaZero has a distributed implementation with multiple actors. We ran tabu search for seven days to make a fair comparison to AlphaZero. Crucially, we ran all the methods until their results plateaued, and the cross-entropy method was run for large time frames for small sizes until no improvement was observed.

\section{Discussion}
We studied a challenging learning-to-search benchmark inspired by an open problem in extremal graph theory, compared a neural network-guided MCTS with tabu search, and observed that using a curriculum is crucial for improving the state of the art, but introducing learning did not yield improvements (a similar phenomenon was observed for another extremal graph theory problem~\cite{ramseymultiplicity}).
This could be because the problem has lots of local optima and the search space is hard to explore: for some sizes, there is only one feasible graph with the optimal score~\cite{backelin}, and during our experiments, for size $n=96$, only one of our runs found a score of 411; all other runs found smaller scores. 
Also, in contrast to problems on which RL has improved the state of the art and do not have strong local search baselines (e.g., the tensor decomposition problem~\cite{alphatensor}), for this problem, natural, fast, and strong local search algorithms exist.
Finally, in contrast to typical RL problems, where the goal is to maximize the \emph{expected return} in a non-deterministic environment, here we want to maximize the \emph{best-case} return in a deterministic environment---i.e., we need only find a good solution once. So, it's unclear whether the classical RL objective is the right approach here; finding better objectives for learning-to-search is an open research problem.

Some of the ideas in this work---the curriculum, the edge-flipping environment, the novel representation Pairformer, and incremental local search---could be used in similar problems. In particular, our edge-flipping environment allows a flexible curriculum approach, where the high-scoring graph of the smaller size need not be an exact subgraph of the optimal larger graphs. This could prove useful in other mathematical problems that have a similar structure.

Finally, while we have brought the state-of-the-art ML algorithm AlphaZero on par with tabu search, the main improvement came from using an incremental approach. Hence one may ask: What improvements to ML approaches are required to outpace classical heuristics on search problems? 

\clearpage
\section*{Contribution Statement} 
Abbas Mehrabian, Ankit Anand, and Hyunjik Kim contributed equally as joint first authors to the paper. Doina Precup and Adam Zsolt Wagner contributed equally as senior authors and are listed in alphabetical order. Tudor Berariu was doing an internship at Google DeepMind while this work was done.

\section*{Acknowledgements} 
We are grateful to Eser Ayg{\"u}n for advising us throughout this project and his helpful comments on the first draft of the paper. We also thank Brendan McKay for releasing the best-known graphs up to size 64, which helped us jump-start our incremental tabu search algorithm.

We appreciate several useful discussions with
David Applegate, 
Charles Blundell, 
Alex Davies, 
Matthew Fahrbach, 
Alhussein Fawzi,
Michael Figurnov,
David Garnick,
Xavier Glorot,
Harris Kwong,
Felix Lazebnik,
Shibl Mourad, 
Sébastien Racaniere, 
Tara Thomas, and
Theophane Weber.

Joonkyung Lee is supported by Samsung STF Grant SSTF-BA2201-02.

\bibliographystyle{named}
\bibliography{main}

\clearpage

\appendix

\begin{table*}
\setlength{\tabcolsep}{20pt}
\renewcommand{\arraystretch}{2}
\caption{The values of $f(n)$ for $n=1, 2, \dots, 9$.}    
\label{small_fn}
    \centering\begin{tabular}{|c|c|c|}
        \hline
         $n$ & $f(n)$ & all $n$-node feasible graphs with $f(n)$ edges \\
        \hline
         1 & 0 & 
\begin{tikzpicture}
\filldraw [blue] (0,0) circle (2pt);
\end{tikzpicture}
\\ \hline
         2 & 1 & 
\begin{tikzpicture}
\filldraw [blue] (0,0) circle (2pt);
\filldraw [blue] (1,0) circle (2pt);
\draw [blue, thick] (0,0) -- (1,0);
\end{tikzpicture}
         \\  \hline
         3 & 2 & 
\begin{tikzpicture}
\filldraw [blue] (0,0) circle (2pt);
\filldraw [blue] (1,0) circle (2pt);
\filldraw [blue] (2,0) circle (2pt);
\draw [blue, thick] (0,0) -- (2,0);
\end{tikzpicture}
         \\  \hline
         4 & 3 & 
\begin{tikzpicture}
\filldraw [white] (0,1.5) circle (2pt);

\filldraw [blue] (0,0) circle (2pt);
\filldraw [blue] (1,0) circle (2pt);
\filldraw [blue] (0,1) circle (2pt);
\filldraw [blue] (1,1) circle (2pt);
\draw [blue, thick] (0,1) -- (0,0) -- (1,0) -- (1,1);

\filldraw [blue] (4,0) circle (2pt);
\filldraw [blue] (5,0) circle (2pt);
\filldraw [blue] (6,0) circle (2pt);
\filldraw [blue] (5,1) circle (2pt);
\draw [blue, thick] (4,0) -- (6,0) -- (5,0) -- (5,1);
\end{tikzpicture}
         \\  \hline
         5 & 5 & 
\begin{tikzpicture}
\filldraw [white] (0,2) circle (2pt);

\filldraw [blue] (0,0.2) circle (2pt);
\filldraw [blue] (1,0.2) circle (2pt);
\filldraw [blue] (0,1) circle (2pt);
\filldraw [blue] (1,1) circle (2pt);
\filldraw [blue] (0.5,1.5) circle (2pt);
\draw [blue, thick] (0,1) -- (0,0.2) -- (1,0.2) -- (1,1) -- (0.5,1.5) -- (0,1);
\end{tikzpicture}
         \\  \hline
         6 & 6 & 
\begin{tikzpicture}
\filldraw [white] (0,2) circle (2pt);

\filldraw [blue] (0,0) circle (2pt);
\filldraw [blue] (1,0) circle (2pt);
\filldraw [blue] (0,0.8) circle (2pt);
\filldraw [blue] (1,0.8) circle (2pt);
\filldraw [blue] (0.5,1.3) circle (2pt);
\filldraw [blue] (0.5,0.55) circle (2pt);
\draw [blue, thick] (0,0.8) -- (0,0) -- (1,0) -- (1,0.8) -- (0.5,1.3) --(0.5,0.55) -- (0.5, 1.3) -- (0,0.8);

\filldraw [blue] (4,0) circle (2pt);
\filldraw [blue] (5,0) circle (2pt);
\filldraw [blue] (5.5,0.5) circle (2pt);
\filldraw [blue] (5,1) circle (2pt);
\filldraw [blue] (4,1) circle (2pt);
\filldraw [blue] (3.5,0.5) circle (2pt);
\draw [blue, thick] (4,0) -- (5,0) -- (5.5,0.5) -- (5,1) -- (4,1) -- (3.5,0.5) -- (4,0);
\end{tikzpicture}
         \\  \hline
         7 & 8 & 
\begin{tikzpicture}
\filldraw [white] (2,2.5) circle (2pt);

\filldraw [blue] (0,0) circle (2pt);
\filldraw [blue] (1,0) circle (2pt);
\filldraw [blue] (2,0) circle (2pt);
\filldraw [blue] (3,0) circle (2pt);
\filldraw [blue] (4,0) circle (2pt);
\filldraw [blue] (2,1) circle (2pt);
\filldraw [blue] (2,2) circle (2pt);
\draw [blue, thick] (0,0) -- (4,0);
\draw [blue, thick] (2,0) -- (2,2);
\draw [blue, thick] (0,0) -- (2,2);
\draw [blue, thick] (4,0) -- (2,2);
\end{tikzpicture}
         \\  \hline
         8 & 10 & 
\begin{tikzpicture}
\filldraw [white] (2,2.5) circle (2pt);

\filldraw [blue] (0,0) circle (2pt);
\filldraw [blue] (1,0) circle (2pt);
\filldraw [blue] (2,0) circle (2pt);
\filldraw [blue] (3,0) circle (2pt);
\filldraw [blue] (4,0) circle (2pt);
\filldraw [blue] (2,1) circle (2pt);
\filldraw [blue] (2,2) circle (2pt);
\filldraw [blue] (1.5,-1) circle (2pt);
\draw [blue, thick] (0,0) -- (1.5,-1) -- (3,0);
\draw [blue, thick] (0,0) -- (4,0);
\draw [blue, thick] (2,0) -- (2,2);
\draw [blue, thick] (0,0) -- (2,2);
\draw [blue, thick] (4,0) -- (2,2);
\end{tikzpicture}
         \\  \hline
         9 & 12 & 
\begin{tikzpicture}
\filldraw [white] (2,2.5) circle (2pt);

\filldraw [blue] (0,0) circle (2pt);
\filldraw [blue] (1,0) circle (2pt);
\filldraw [blue] (2,0) circle (2pt);
\filldraw [blue] (3,0) circle (2pt);
\filldraw [blue] (4,0) circle (2pt);
\filldraw [blue] (2,1) circle (2pt);
\filldraw [blue] (2,2) circle (2pt);
\filldraw [blue] (1.5,-1) circle (2pt);
\filldraw [blue] (2.5,-1) circle (2pt);
\draw [blue, thick] (1,0) -- (2.5,-1) -- (4,0);
\draw [blue, thick] (0,0) -- (1.5,-1) -- (3,0);
\draw [blue, thick] (0,0) -- (4,0);
\draw [blue, thick] (2,0) -- (2,2);
\draw [blue, thick] (0,0) -- (2,2);
\draw [blue, thick] (4,0) -- (2,2);
\end{tikzpicture}\\  \hline
    \end{tabular}
\end{table*}

\clearpage

\begin{table}
\caption{The maximum number of edges of graphs with no 3-cycles or 4-cycles found by incremental tabu search, and the number of non-isomorphic graphs found for each size.}
\label{table:its}
\begin{tabular}{rrr}
\toprule
 Num. nodes &  Num. edges &  Num. graphs \\
\midrule
         1 &          0 &           1 \\
         2 &          1 &           1 \\
         3 &          2 &           1 \\
         4 &          3 &           2 \\
         5 &          5 &           1 \\
         6 &          6 &           2 \\
         7 &          8 &           1 \\
         8 &         10 &           1 \\
         9 &         12 &           1 \\
        10 &         15 &           1 \\
        11 &         16 &           3 \\
        12 &         18 &           7 \\
        13 &         21 &           1 \\
        14 &         23 &           4 \\
        15 &         26 &           1 \\
        16 &         28 &          22 \\
        17 &         31 &          14 \\
        18 &         34 &          15 \\
        19 &         38 &           1 \\
        20 &         41 &           1 \\
        21 &         44 &           2 \\
        22 &         47 &           3 \\
        23 &         50 &           7 \\
        24 &         54 &           1 \\
        25 &         57 &           6 \\
        26 &         61 &           2 \\
        27 &         65 &           1 \\
        28 &         68 &           4 \\
        29 &         72 &           1 \\
        30 &         76 &           1 \\
        31 &         80 &           2 \\
        32 &         85 &           1 \\
        33 &         87 &          12 \\
        34 &         90 &         230 \\
        35 &         95 &           5 \\
        36 &         99 &          34 \\
        37 &        104 &           6 \\
        38 &        109 &           2 \\
        39 &        114 &           1 \\
        40 &        120 &           1 \\
        41 &        124 &           1 \\
        42 &        129 &           1 \\
        43 &        134 &           1 \\
        44 &        139 &           2 \\
        45 &        145 &           1 \\
        46 &        150 &           2 \\
        47 &        156 &           1 \\
        48 &        162 &           1 \\
        49 &        168 &           1 \\
        50 &        175 &           1 \\
\bottomrule
\end{tabular}
\end{table}

\begin{table}
\begin{tabular}{rrr}
\toprule
 Num. nodes &  Num. edges &  Num. graphs \\
\midrule
        51 &        176 &           7 \\
        52 &        178 &         102 \\
        53 &        181 &         402 \\
        54 &        185 &           5 \\
        55 &        189 &           5 \\
        56 &        193 &          11 \\
        57 &        197 &           4 \\
        58 &        202 &           1 \\
        59 &        207 &           2 \\
        60 &        212 &           2 \\
        61 &        216 &          16 \\
        62 &        220 &          85 \\
        63 &        224 &        2662 \\
        64 &        230 &           2 \\
        65 &        235 &          74 \\
        66 &        241 &           1 \\
        67 &        246 &          43 \\
        68 &        251 &         979 \\
        69 &        257 &          85 \\
        70 &        262 &        1575 \\
        71 &        268 &          66 \\
        72 &        273 &         694 \\
        73 &        279 &           4 \\
        74 &        284 &         172 \\
        75 &        290 &          12 \\
        76 &        295 &        1298 \\
        77 &        301 &           1 \\
        78 &        306 &         548 \\
        79 &        312 &          39 \\
        80 &        318 &          25 \\
        81 &        324 &          11 \\
        82 &        329 &         673 \\
        83 &        335 &          22 \\
        84 &        340 &         375 \\
        85 &        346 &           9 \\
        86 &        352 &           4 \\
        87 &        357 &         584 \\
        88 &        363 &         288 \\
        89 &        369 &          50 \\
        90 &        375 &           8 \\
        91 &        381 &           2 \\
        92 &        387 &          10 \\
        93 &        393 &           1 \\
        94 &        398 &        1014 \\
        95 &        404 &         605 \\
        96 &        411 &           1 \\
        97 &        417 &           4 \\
        98 &        422 &         819 \\
        99 &        428 &         161 \\
       100 &        434 &          49 \\
\bottomrule
\end{tabular}

\end{table}

\begin{table}

\begin{tabular}{rrr}
\toprule
 Num. nodes &  Num. edges &  Num. graphs \\
\midrule
         101 &        440 &          15 \\
       102 &        446 &           3 \\
       103 &        452 &           5 \\
       104 &        458 &           8 \\
       105 &        464 &          10 \\
       106 &        470 &           6 \\
       107 &        476 &           9 \\
       108 &        482 &           7 \\
       109 &        488 &          14 \\
       110 &        494 &         100 \\
       111 &        500 &         119 \\
       112 &        506 &         105 \\
       113 &        513 &           4 \\
       114 &        519 &          30 \\
       115 &        526 &           1 \\
       116 &        532 &           3 \\
       117 &        538 &          34 \\
       118 &        544 &          27 \\
       119 &        551 &           5 \\
       120 &        557 &          33 \\
       121 &        564 &           1 \\
       122 &        570 &          22 \\
       123 &        577 &           1 \\
       124 &        583 &          18 \\
       125 &        589 &          67 \\
       126 &        596 &           3 \\
       127 &        603 &           1 \\
       128 &        609 &           3 \\
       129 &        616 &          34 \\
       130 &        623 &          15 \\
       131 &        630 &           1 \\
       132 &        636 &          41 \\
       133 &        643 &           6 \\
       134 &        649 &          40 \\
       135 &        656 &           3 \\
       136 &        663 &           1 \\
       137 &        669 &          54 \\
       138 &        676 &          32 \\
       139 &        683 &          17 \\
       140 &        690 &          18 \\
       141 &        697 &           9 \\
       142 &        704 &           7 \\
       143 &        711 &           4 \\
       144 &        718 &           3 \\
       145 &        725 &           3 \\
       146 &        732 &          12 \\
       147 &        739 &           4 \\
       148 &        746 &           2 \\
       149 &        752 &           9 \\
       150 &        759 &           7 \\
\bottomrule
\end{tabular}
\end{table}

\begin{table}
\begin{tabular}{rrr}
\toprule
 Num. nodes &  Num. edges &  Num. graphs \\
\midrule
         151 &        766 &           8 \\
       152 &        773 &           5 \\
       153 &        780 &           2 \\
       154 &        788 &           3 \\
       155 &        795 &          14 \\
       156 &        802 &           1 \\
       157 &        809 &          10 \\
       158 &        816 &          18 \\
       159 &        823 &          27 \\
       160 &        830 &          12 \\
       161 &        838 &           2 \\
       162 &        845 &          16 \\
       163 &        852 &          24 \\
       164 &        860 &           7 \\
       165 &        867 &           5 \\
       166 &        874 &           4 \\
       167 &        881 &          20 \\
       168 &        888 &          32 \\
       169 &        896 &           3 \\
       170 &        904 &           4 \\
       171 &        911 &          19 \\
       172 &        919 &           1 \\
       173 &        926 &           7 \\
       174 &        933 &          21 \\
       175 &        940 &           4 \\
       176 &        947 &           4 \\
       177 &        954 &          46 \\
       178 &        962 &           4 \\
       179 &        970 &           1 \\
       180 &        977 &           2 \\
       181 &        984 &          12 \\
       182 &        992 &           6 \\
       183 &       1000 &           8 \\
       184 &       1008 &           6 \\
       185 &       1015 &           1 \\
       186 &       1022 &           2 \\
       187 &       1024 &          13 \\
       188 &       1034 &           1 \\
       189 &       1044 &           1 \\
       190 &       1050 &           1 \\
       191 &       1056 &           1 \\
       192 &       1065 &           1 \\
       193 &       1069 &           4 \\
       194 &       1070 &           1 \\
       195 &       1082 &           1 \\
       196 &       1069 &           5 \\
       197 &       1079 &           1 \\
       198 &       1086 &           7 \\
       199 &       1094 &           6 \\
       200 &       1096 &           2 \\
\bottomrule
\end{tabular}

\end{table}

\clearpage

\section{An example}
\label{sec:example}
As a concrete example of our results, we provide the sparse6 representation~\cite{sparse6} of one graph on 64 nodes with 230 edges without 3- or 4-cycles we found via incremental tabu search (the best published bound is 229~\cite{mckay_graphs}):
{\scriptsize
\begin{samepage}
\begin{verbatim}
b'>>sparse6<<:~?@?_OGoIO?AoxKSKFDO@SQKKbBXCGECAo`KwdOj?_|E
lDHeGOgrKFoOgcYMjFH?cpLLFh`[y^_Bq[OWQIeGwwaTNOpX`?`ba@xH@D
HEr{WQSIeSCCUONghp?mWRQ@OseTJDw_keSLfRs?GFRiWP?iZMHGWWO\\P
OPO{aSKP@?wsc_ERiMEZSIdcSWU``p_s_XbFeRMMSJebaDCtecaaLGrhft
at[nXtRiX]oYLWH[qZVLw_c_UOmhG{mgWKU[CETLHTIlfCDHTjDjBFDAp|
|\n'
\end{verbatim}
\end{samepage}}

\section{Proof of Lemma~\ref{lemma:score}}\label{sec:proof}
Let $G$ be any $n$-node graph and apply the following procedure: while $G$ has at least one 3-cycle or 4-cycle, choose any such cycle arbitrarily and delete one of its edges; repeat until no 3- or 4-cycles remain. Denote the resulting graph by $G'$.
Since $\triangle(G')=\square(G')=0$, we have $\score(G') = e(G') \leq f(n)$.
Also, for each 3-cycle or 4-cycle of $G$, we have deleted at most one edge during this procedure, hence $s(G') = e(G') \geq e(G) - \triangle(G) - \square(G) = s(G)$.
Therefore, for all $n$-node graphs $G$, $s(G) \leq s(G') \leq f(n)$, proving the first part of the lemma.
The second part of the lemma follows from the definition of $f(n)$: see~\eqref{def_fn}.

\section{Hyperparameters}\label{sec:hyperparams}
The hyperparameters of cross-entropy method \cite{wagner_paper} is described in table \ref{table:ce_hyperparameters}. The hyperparameters of Alpha-Zero is described in table \ref{table:az_hyperparameters} 

\begin{table}
\centering
\caption{The hyperparameters for AlphaZero agent.}
\begin{tabular}{ll}
     \toprule
\textbf{Hyperparamter}             & \textbf{Value} \\
\midrule
Max training steps                 & 1e6            \\
Total batch size                   & 8              \\
ResNet NumberOfChannels            & 256            \\
ResNet NumberOfLayers              & 10             \\
ResNetNumberOfHeads                & 8              \\
ResNet HeadDepth                   & 128            \\
Pairformer NumberOfLayers          & 3              \\
Pairformer NumberOfHeads           & 8              \\
PairformeNumberOfChannels          & 64             \\
Pairformer UseLayerNorm            & True           \\
WeightDecay                        & 1e-5           \\
TargetNetworkUpdateInterval(value) & 100            \\
SimulationsPerMove                 & 400            \\
PolicyInferenceSamples             & 32             \\
PolicyTrainingSamples              & 32             \\
Initial Learning Rate(LR)          & 3e-3           \\
LR Decay factor                    & 0.1            \\
LR decay steps                     & 5e5           
\end{tabular}
\label{table:az_hyperparameters}
\end{table}

\begin{table}
\centering
\caption{The hyperparameter sweep for the cross-entropy method. ``Examples per iteration'' refers to the number of proposed constructions to sample on each iteration before sampling the top k percent and training the neural network on those highest scoring examples.}
\begin{tabular}{c c c}
     \toprule
     Learning rate & Examples per iteration & Top k \\
     \midrule
     1e-4, 1e-5 & 100, 1000 & 5, 8, 10
\end{tabular}
\label{table:ce_hyperparameters}
\end{table}

\end{document}